\pdfoutput=1

\documentclass[a4paper, 10pt, conference]{ieeeconf}      

\usepackage{blindtext, graphicx}
\usepackage{gensymb}
\usepackage{xcolor}

\usepackage{tikz,lipsum}
\usepackage[labelformat=simple]{subcaption}

\DeclareCaptionLabelSeparator{periodspace}{.\quad}
\usepackage{listings}

\usepackage{algorithm}
\usepackage{algorithmic}

\usepackage{ amssymb }
\usepackage{amsmath}
\usepackage{commath}

\usepackage{framed} 

\usepackage{float}

\newcommand*{\prob}{\mathsf{P}}

\IEEEoverridecommandlockouts                              

\title{\LARGE \bf
Multi-Objective Convolutional Neural Networks for Robot Localisation and 3D Position Estimation in 2D Camera Images
}

\author{Justinas Mi\v{s}eikis$^{1}$, Inka Brijacak$^{2}$, Saeed Yahyanejad$^{3}$, Kyrre Glette$^{4}$, Ole Jakob Elle$^{5}$, Jim Torresen$^{6}$
\thanks{$^{1}$ $^{4}$ $^{5}$ $^{6}$Justinas Mi\v{s}eikis, Kyrre Glette, Ole Jakob Elle and Jim Torresen are with the Department of Informatics, University of Oslo, Oslo, Norway}
\thanks{$^{2}$ $^{3}$ Inka Brijacak and Saeed Yahyanejad are with the Joanneum Research - Robotics, Klagenfurt am W\"orthersee, Austria} 
\thanks{$^{5}$Ole Jakob Elle has his main affiliation with The Intervention Centre, Oslo University Hospital, Oslo, Norway {\tt\small oelle@ous-hf.no}}%
\thanks{$^{1}$ $^{4}$ $^{6}$ {\tt\small \{justinm,kyrrehg,jimtoer\}@ifi.uio.no}}%
\thanks{$^{2}$ {\tt\small Inka.Brijacak@joanneum.at}}%
\thanks{$^{3}$ {\tt\small Saeed.Yahyanejad@joanneum.at}}%
}

\begin{document}

\maketitle
\thispagestyle{empty}
\pagestyle{empty}

\begin{abstract}

The field of collaborative robotics and human-robot interaction often focuses on the prediction of human behaviour, while assuming the information about the robot setup and configuration being known. This is often the case with fixed setups, which have all the sensors fixed and calibrated in relation to the rest of the system. However, it becomes a limiting factor when the system needs to be reconfigured or moved. We present a deep learning approach, which aims to solve this issue. Our method learns to identify and precisely localise the robot in 2D camera images, so having a fixed setup is no longer a requirement and a camera can be moved. In addition, our approach identifies the robot type and estimates the 3D position of the robot base in the camera image as well as 3D positions of each of the robot joints. Learning is done by using a multi-objective convolutional neural network with four previously mentioned objectives simultaneously using a combined loss function. The multi-objective approach makes the system more flexible and efficient by reusing some of the same features and diversifying for each objective in lower layers. A fully trained system shows promising results in providing an accurate mask of where the robot is located and an estimate of its base and joint positions in 3D. We compare the results to our previous approach of using cascaded convolutional neural networks. 

\end{abstract}



\section{INTRODUCTION}

With the tendency of robotic hardware becoming cheaper and more powerful, robots are entering our everyday environments. Household robots like vacuum cleaners do not surprise people anymore. Even faster robot adoption happens in hospitals, warehouses and factories. An important reason for this is advancements in environment perception capabilities. Instead of fencing off the robots, the concept of Industry 4.0 is aimed at having a new era of collaborative robots, which are safe to operate in shared workspaces with humans~\cite{lee2015cyber}. The concept of a shared workspace has been an active research area for many years, which is still highly relevant today~\cite{roach1987coordinating}~\cite{leitner2012transferring}. The industry is catching up to research with robotic platforms like Baxter and Sawyer, which are known to be fully safe to operate around humans. However, they are still at a stage, where collision detection is the main safety system~\cite{fitzgerald2013developing}. But we are looking at more sensitive environments, for example, hospitals, where collision detection is not good enough and full avoidance is needed.

One of the most common methods to observe the environment is by using vision sensors. In this application, 3D cameras observe the workspace and indicate the areas, which are free of obstructions and are safe to operate in as well as obstacles, which should be avoided. Given this information, a robot can find the safest path to reach its goal. However, normally these sensors are fixed either in relation to the robot or in the environment. In order to function in the same coordinate frame and provide accurate information to the robotic system, Hand-Eye calibration is performed. It works well as long as the setup of the sensors and the robot base stays static. If any of them are moved, intentionally or accidentally, the calibration has to be repeated in order for the sensors to work with the necessary precision. Despite some automatic calibration procedures, the process can still be time-consuming, and the system has to be halted until this issue is resolved~\cite{miseikis2016automatic}. 

\begin{figure*}[ht]
\vspace{0.2cm}
\centering
    \includegraphics[width=0.99\linewidth]{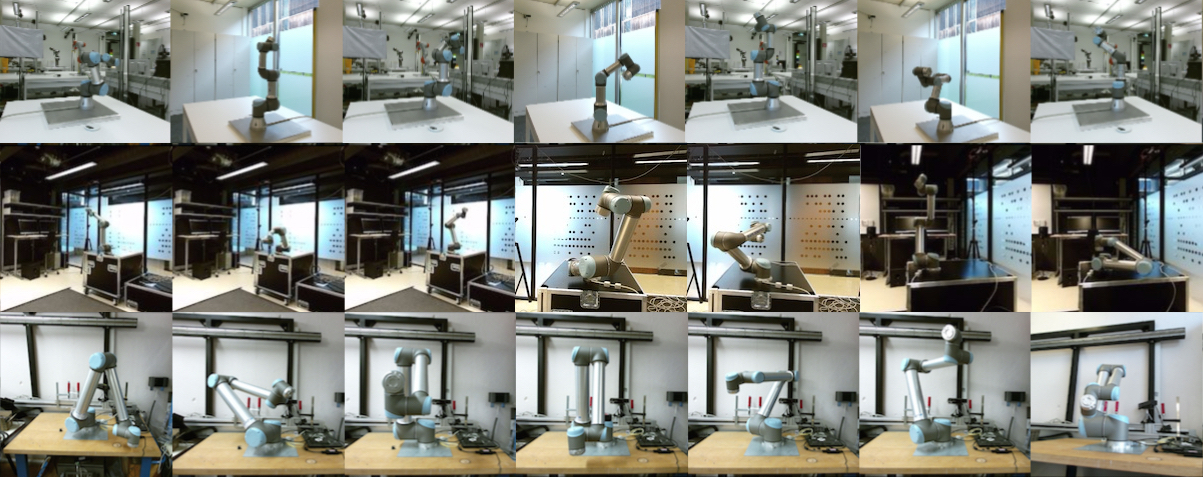}
    \caption{Samples from a collected robot dataset. Each row of images represents different robot type in the following order: UR3, UR5 and UR10. The dataset was created using a varying background to provide more robustness.}
    \label{fig:dataset_example_images}
\vspace{-0.5cm}
\end{figure*}



One way to make the environment aware robots is to use long-term environment observation. Such approaches have been used in the development of robot autonomy and self-localisation tasks. This is commonly developed as navigation algorithms for mobile robot platforms to find their way around in the environment and avoid any static or dynamic obstacles on the way. Typically, robot model and dynamics are typically known~\cite{schneegans2007using}~\cite{gutmann1996amos}~\cite{stasse2008real}~\cite{de2007skeleton}. 

Visual-based robot manipulator tracking has been extensively researched as well. End-effector being the main point of focus with the aim of conducting robot control based on visual servoing~\cite{wilson1996relative}~\cite{ruf1997visual}. Furthermore, it has proven to be an effective method for adaptive redundant robot control in Cartesian space~\cite{daachi2006neural}. Image-based tracking of 7-DoF robot arm showed promising results with dynamic parameter tuning as well~\cite{siradjuddin2014image}. In another project, authors use particle swarm optimisation method for fuzzy sliding mode control to track the end-effector of the robot manipulator~\cite{soltanpour2013particle}. Furthermore, robotic arms were combined with deep learning approaches to learn direct motor commands by using visual inputs. They were based on reinforcement learning and by trying thousands of grasps reaching impressive results of adaptive grasping approaches. However, that required many hours of training while using real hardware~\cite{pinto2016supersizing}~\cite{gu2017deep}~\cite{levine2016end}.

One thing that majority of discussed systems have in common is that prior knowledge of the robotic platforms is given or the setups in regards to hardware are fixed. Any changes to the setup would require re-calibration or at least fine-tuning the algorithms to achieve the same level of performance. Furthermore, common obstacle avoidance algorithms for robotic arms are focusing on the end-effector instead of the whole robot body.

Having non-fixed setup allows easier camera placement in cluttered environments with multiple robots, like a factory floor or automated surgery room. Normally, there is limited space and equipment might have to be shifted around quite frequently. This results in limited line-of-sight or people standing in front of the sensor. Having a multi-camera setup can add the needed redundancy, or using a wearable camera would provide a viewpoint of the operator. On a factory floor, such a camera-based system can give an indication of all the robots located around the person wearing it. A warning or even an emergency stop option can be incorporated into the system for the situations when the robot gets too close to the person within its field of view to ensure a safe operation.

A similar approach could be also used in robot-robot interaction cases, where similar or heterogeneous robots are working in the same environment. Even without having direct communication channels, robots can avoid collisions with each other. On the other hand, this can be used as a redundant navigation system, given the map of the main robots is known, the mobile platform can re-localise itself according to their detected positions. Collaborative tasks would be targeted also, where robots have to hand over tools or work together. Having an active communication channel is not always reliable, so being able to identify robot arms in the environment and their configuration using on-board camera can allow to solve these problems. Provided high enough processing power, swarm robotics could benefit from such systems, where each individual is making independent decisions without any centralised system.

Our current research targets this problem by trying to add flexibility to the robot identification and having easily adjustable setups. One goal is to have a free moving camera and remove the need for Hand-Eye calibration. Instead of having a known transformation matrix between the coordinate frames of the sensor and the robot base, we teach the system to identify the robot body in a 2D color image provided by the vision sensor. This would allow having cameras placed on moving objects, for example, wearable ones or placed on other robots moving in the environment. Our method uses convolutional neural networks (CNNs), which learn visual cues allowing it to understand the environment~\cite{simard2003best}. The system identifies the robot body in the color image, and depth information normally provided by 3D cameras is not needed for the recognition task anymore. Furthermore, the system estimates the robot body configuration and 3D coordinates of each joint of the robot. 

Current work is an extension and improvement of our previously proposed method to use cascaded CNNs (C-CNN) in order to solve this problem~\cite{2018arXiv180102025M}. The advantage of using multi-objective CNN is the ability to train the network on multiple tasks simultaneously while re-using the same features instead of having to re-learn some of them when using C-CNNs. Similar multi-objective CNN approaches have been used for detecting facial landmarks, face recognition and localisation as well as orientation~\cite{yin2017multi}~\cite{wu2015facial}. Other similar approaches can be done to optimise the training of the network on two GPUs, each one following each branch~\cite{krizhevsky2012imagenet}. Also, mid-layer parameter transfer between two identical networks, but each one having different sets of objective labels has proven to be effective~\cite{oquab2014learning}.

This paper is organized as follows. We present the system setup and dataset collection in Section~\ref{sec:system_setup}. Then, we explain the proposed method and CNN architecture in Section~\ref{sec:method} and the training procedure in Section~\ref{sec:training}. We provide experimental results in Section~\ref{sec:results}, followed by relevant conclusions and future work in Section~\ref{sec:conclusion}.


\section{SYSTEM SETUP AND DATASET COLLECTION}
\label{sec:system_setup}

\begin{figure}[ht]
\centering
\begin{subfigure}[t]{0.23\textwidth}
    \includegraphics[width=\textwidth]{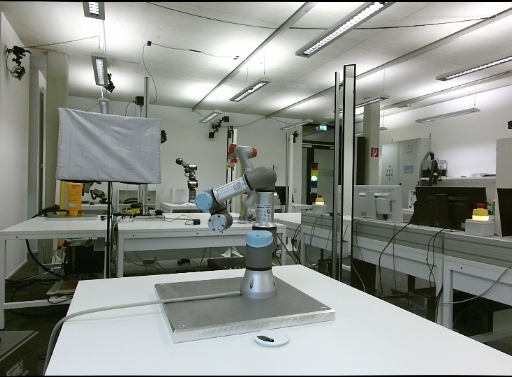}
    \caption{Color image from the dataset used as an input.}
    \label{fig:input_color_image}
\end{subfigure}
~
\begin{subfigure}[t]{0.23\textwidth}
    \includegraphics[width=\textwidth]{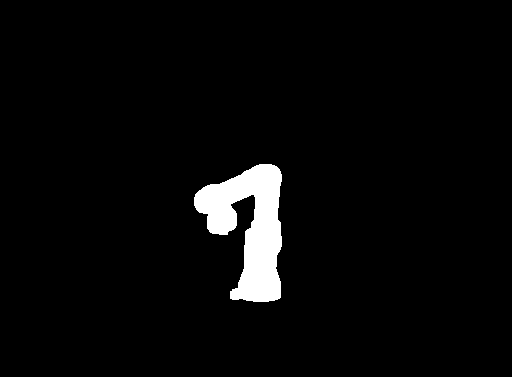}
    \caption{Ground truth model of the robot mask.}
    \label{fig:input_gt_image}
\end{subfigure}
~
\begin{subfigure}[t]{0.23\textwidth}
    \includegraphics[width=\textwidth]{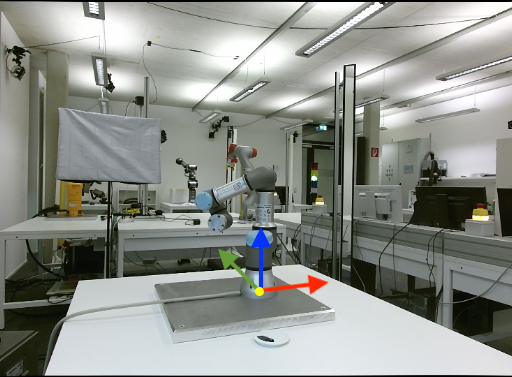}
    \caption{Ground truth data of the robot base 3D position in relation to the camera marked on the input image.}
    \label{fig:input_gt_robot_base}
\end{subfigure}
~
\begin{subfigure}[t]{0.23\textwidth}
    \includegraphics[width=\textwidth]{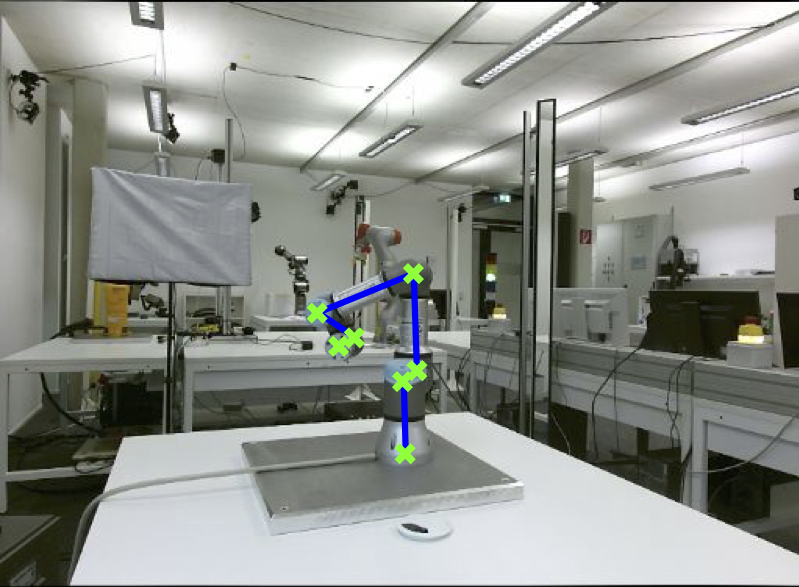}
    \caption{Ground truth data of the 3D position of robot joints marked on the input image.}
    \label{fig:input_gt_joints}
\end{subfigure}
\caption{Example image of the dataset and ground truth examples of the UR3 robot.}
\label{fig:input_data}
\vspace{-0.6cm}
\end{figure}

Deep learning typically requires large amounts of diverse training data for robust learning. However, this is an issue for industrial robotics applications, because there are close to none existing public datasets with well-marked ground truth data. Thus, in order to get reliable training data, a new dataset was created specifically for the presented application. The whole range of Universal Robots: UR3, UR5 and UR10, were used at three institutions: TU Graz, Joanneum Research and the University of Oslo. All three robots share similar visual appearance, but differ significantly in size, reach and payload capabilities.

As a vision sensor, a Kinect V2 camera is used~\cite{Fankhauser2015KinectV2ForMobileRobotNavigation}. It provides both color image and depth information. Depth images are only used for the creation of the ground truth data, while the whole following recognition process is using just a color image as an input.

For each recording, in order to have a precise ground truth data, Kinect was placed at arbitrary position observing the workspace of the robot. At each position, a Hand-Eye calibration was performed by placing a marker on the end-effector of the robot and using both color and depth image for the calibration process~\cite{heikkila2000flexible}. This provides an accurate coordinate frame transformation between the camera and the base of the robot, with an error below $0.52$ cm for all the datasets.

\begin{table}[h]
\caption{Dataset summary describing a number of samples collected for each type of the robot. In total 9 recordings were made, 3 for each type of robot.}
\label{table:dataset_summary}
\centering
\begin{tabular}{ |p{2.0cm}||p{1.5cm}|p{3cm}|}
 \hline
 Recording & Robot Type & Number of Samples \\
 \hline
 Rec 1 & UR3 & 211\\
 Rec 2 & UR3 & 252\\
 Rec 3 & UR3 & 463\\
 \hline
 Rec 4 & UR5 & 252\\
 Rec 5 & UR5 & 756\\
 Rec 6 & UR5 & 1512\\
 \hline
 Rec 7 & UR10 & 112\\
 Rec 8 & UR10 & 278\\
 Rec 9 & UR10 & 514\\
 \hline
\end{tabular}
\vspace{-0.2cm}
\end{table}

Once the transformation is known, a mask defining the location of the robot in the camera image can be calculated. It is done by utilising the encoder information from each joint of the robot and using a simplified model of the robot. The robot is represented using basic cylindrical and spherical shapes in 3D space according to its model and then mapped onto a virtual 2D image representing the sight of the camera. Thresholding this image results in a robot body mask representation in the camera image. The MoveIt! package was used to implement this method~\cite{sucan2013moveit}.

The robot should be observed from all the different angles and in a high variety of joint angle configurations to achieve good robustness. Movements for the data collection were programmed to provide a high diversity of viewpoints and robot body configurations. Each robot joint is moved through the full range of motion in combination with other joints as well. The step size of the joint movements is varied between the datasets resulting in a different number of samples in each. After each movement, a trigger signal is used to save the data. At each instance, camera images, joint coordinates, Cartesian coordinates of each joint and ground-truth robot mask images were saved. The number of samples per dataset varied from $112$ to $1512$. The variation was caused by different types and resolutions of programmed robot movements during the data collection. In total $9$ datasets were collected, $3$ for each type of the robot, summarized in Table~\ref{table:dataset_summary}. Example images from the collected dataset are shown in Figure~\ref{fig:dataset_example_images}. Datasets with UR5 robot were the most extensive given the access to the robot at the lab of the main author. An example of color and ground truth of robot mask, base position and joint positions can be seen in Figure~\ref{fig:input_data}.

\begin{figure*}[ht]
\vspace{0.2cm}
    \centering
    \includegraphics[width=0.90\textwidth]{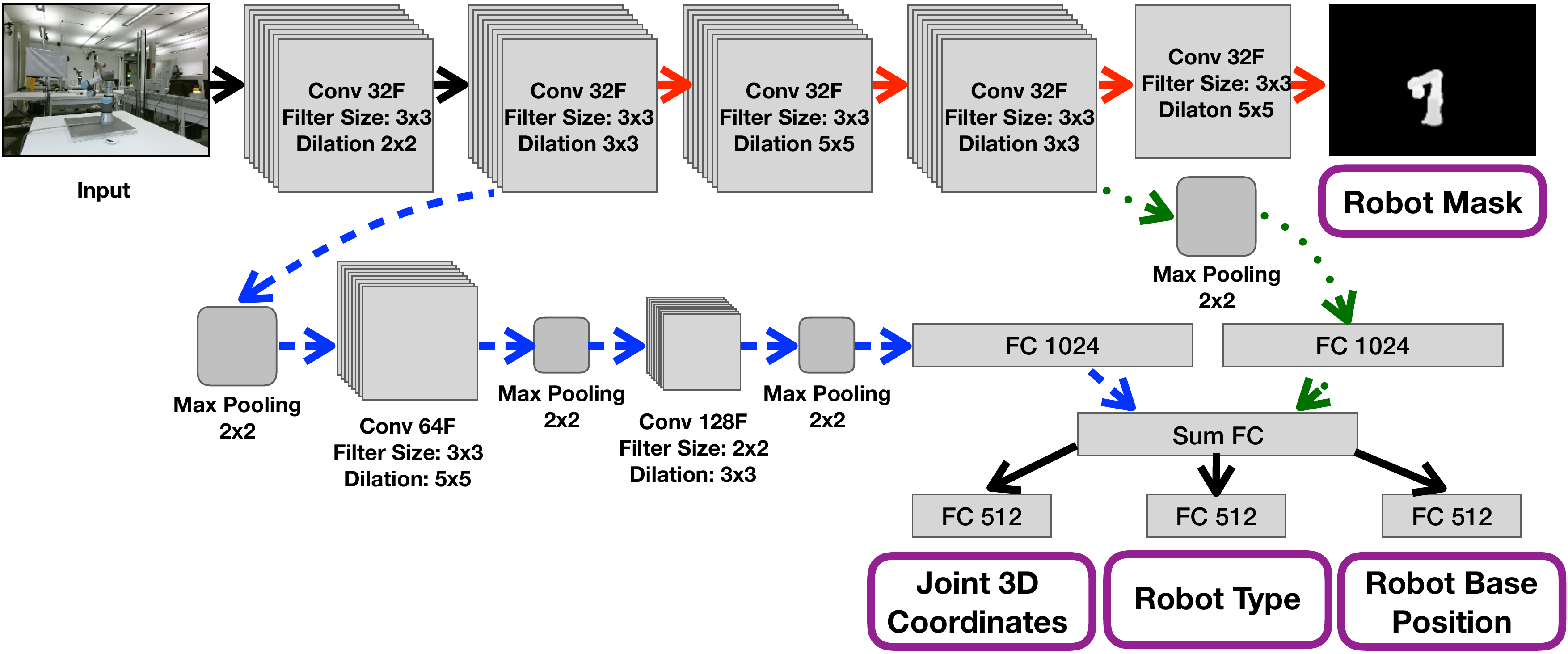}
    \caption{Multi-objective CNN structure. Input is a simple 2D color image and the network is trained for four outputs: robot mask, 3D coordinates of robot joints, 3D coordinates of robot base position and robot type. There are two main branches of the CNN. The first one is aimed to learn the features leading to an accurate robot mask mainly consisting of dilated convolutional layers. It is marked by red solid arrows. The second branch, marked in blue dashed arrows, consists of a number of max pooling and dilated convolutional layers with fully connected layers at the end. The goal is to predict coordinates as a regression task as well as classify the robot type. Additionally, there is a branch starting from the 4th convolutional layer of robot mask task to the end of the blue branch using summing of fully connected layers, marked in dotted green arrows. It adds the information of features well defining the visual representation of the robot to the other tasks further improving the results. The whole CNN is trained for all four outputs simultaneously using a common loss function.}
    \label{fig:multi_obj_cnn_structure}
\vspace{-0.6cm}
\end{figure*}

Recorded images have $512\times424$ pixel resolution and they are all rectified to compensate for lens distortion. Internal camera calibration was used to ensure that both color and depth information have a good overlap, avoiding any offsets. Random sampling was used to divide the final dataset into the training set and the test set by ratios of $80\%$ and $20\%$ of all the images respectively.

\section{METHOD}
\label{sec:method}

Our approach is based on a multi-objective CNN structure. This approach allows us to get multiple outputs of different types by having just one input. It is achieved by having a number of convolutional layers, which are common for the whole system and then branching out the structure for each of the objectives. The whole system is trained simultaneously, meaning that the features in common layers are reused.

In our case, we train for four objectives:
\begin{itemize}
  \item Robot mask in the image
  \item Robot type
  \item 3D Robot base position in relation to the camera
  \item 3D Position of the robot joints
\end{itemize}

The structure of the CNN is shown in Figure~\ref{fig:multi_obj_cnn_structure}. It consists of the two main branches. The first one learns a classification task of finding the robot in the input image. It results in a robot mask defining the location of the robot. The second branch is for the regression tasks of finding the 3D robot base coordinates in relation to the camera and the 3D coordinates of each of the robot joints. In addition, on the same branch, the classification of the robot type is done.

In addition, there is the second branch from the 4th convolutional layer towards the robot mask, which connects to the second branch. Given the idea that robot body parts are learned quite well for the robot mask classification task, this additional input provides the essential information for identifying the location of the robot joints. Fully connected layers, which are summed, are believed to filter the important visual cues and assist for the coordinate regression tasks.

\subsection{Loss Functions}


Loss functions are used to determine the quality of training. Given we have four objectives, we first describe loss functions for each one. Because the network is trained for all of the objectives simultaneously, finally we combine all four loss function into one used for the actual training.


The loss function for the robot mask was designed to adjust for a small area the foreground object takes up in the input image. In our datasets, the area taken up by the robot body in the input image was varying between $6-17\%$ of the whole image. If the loss function does not compensate for this, the CNN could classify all the pixels as background and still achieve the accuracy of $83$ to $94\%$, which is conceptually wrong. To prevent this, the foreground weight $w_{fg}$ is calculated, as described in Equation~\ref{eq:fg_weight}. It is based on the inverse probability of the foreground and background classes, where $Y \in \{fg, bg\}$.


\begin{equation}
    w_{fg} = \frac{1}{\prob(Y=fg)}
\label{eq:fg_weight}
\end{equation}

The background weight $w_{bg}$ is calculated in Equation~\ref{eq:bg_weight}.

\begin{equation}
    w_{bg} = \frac{1}{\prob(Y=bg)}
\label{eq:bg_weight}
\end{equation}

\begin{figure*}[ht]
\vspace{0.2cm}
    \hfill
    \centering
    \begin{subfigure}[t]{0.32\textwidth}
        \centering
        \includegraphics[width=\linewidth]{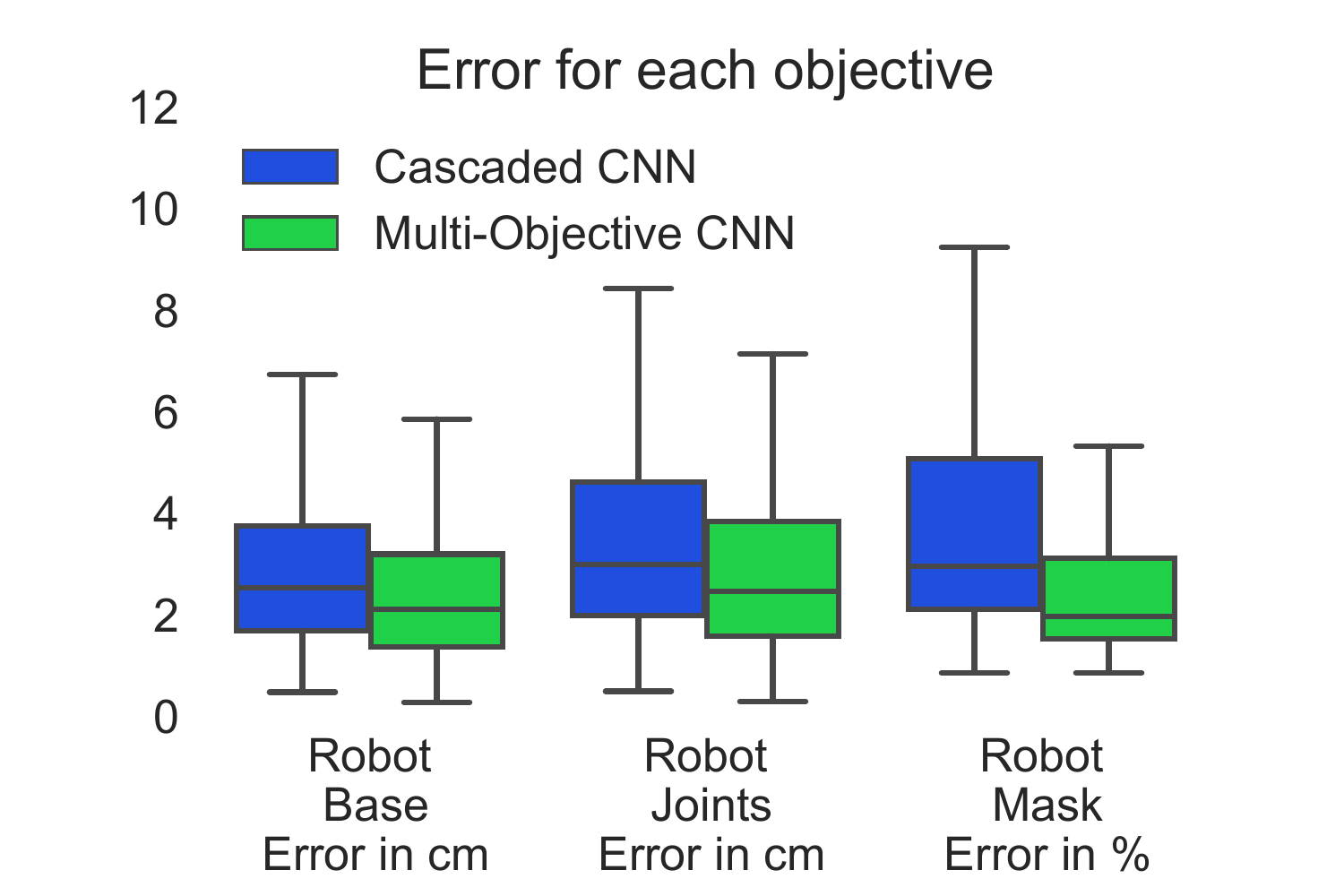}
        \caption{Comparison of the Multi-Objective CNN approach against the Cascaded CNN approach~\cite{2018arXiv180102025M}. Errors for all the objectives are smaller, as well as the range of quartile values.}
        \label{fig:results_all} 
    \end{subfigure}
    \hfill
    \begin{subfigure}[t]{0.32\textwidth}
        \centering
        \includegraphics[width=\linewidth]{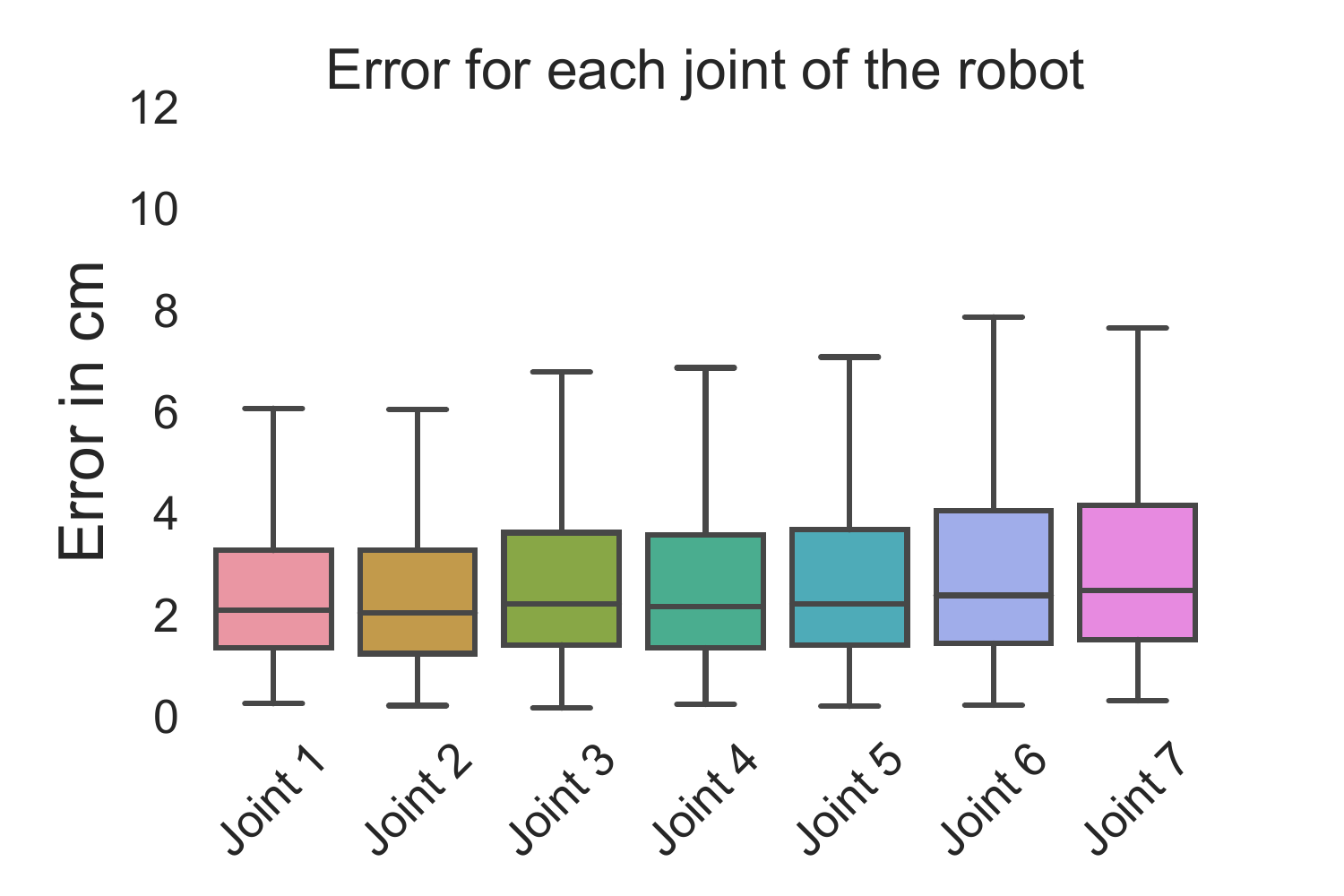}
        \caption{Error for the 3D coordinate estimation for positions of each robot joint. It can be seen that the error slightly increases for joints further away from the base.}
        \label{fig:results_joints}
    \end{subfigure}
    \hfill
    \begin{subfigure}[t]{0.32\textwidth}
        \centering
        \includegraphics[width=\linewidth]{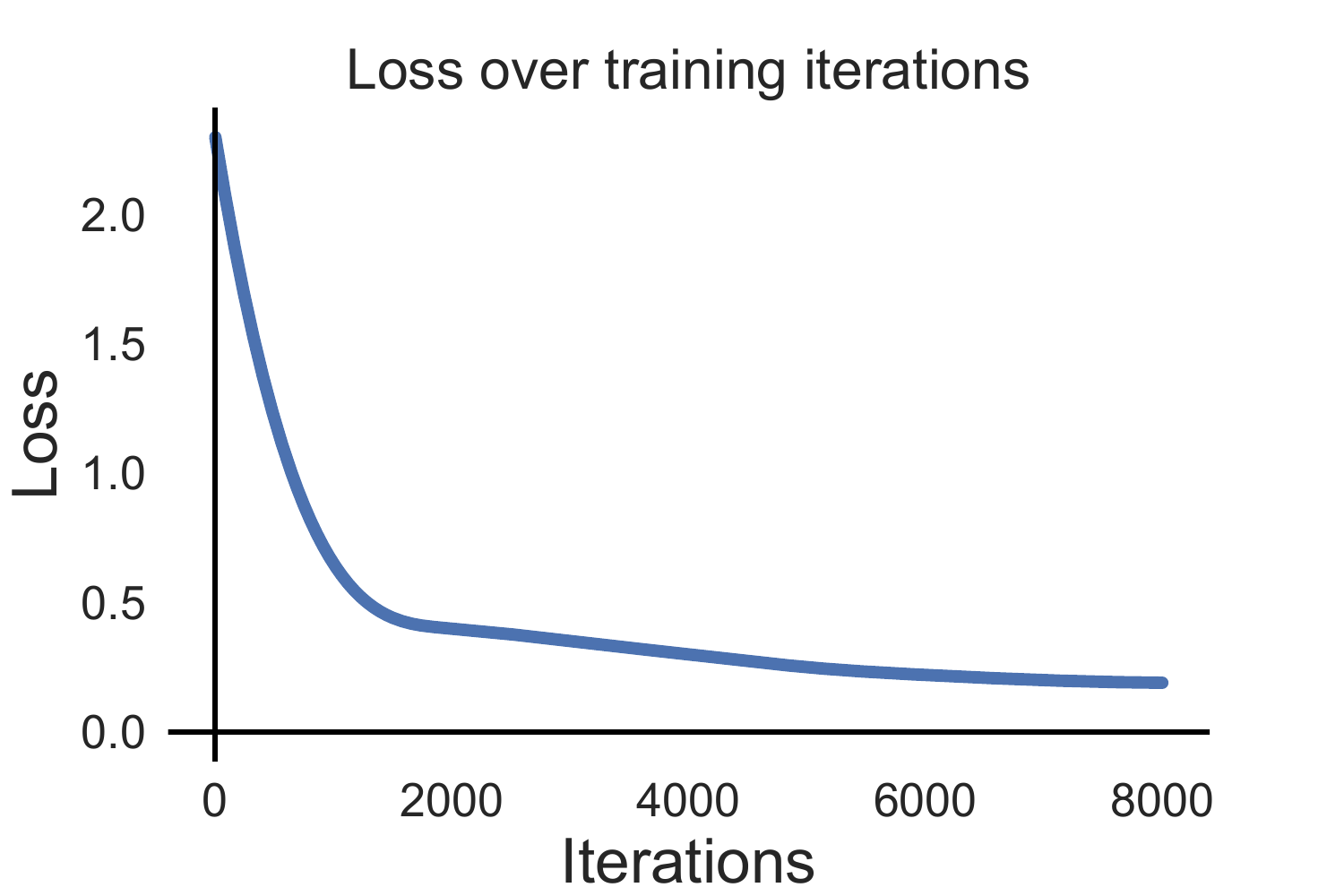}
        \caption{Value of the loss function on the test set over iterations during the training process.}
        \label{fig:results_loss}
    \end{subfigure}
    
    \caption{Evaluation of our method by testing the trained system on the test dataset.}
\vspace{-0.6cm}
\end{figure*}

The robot mask loss function is calculated in two steps. First, a per-pixel loss $l^n$ is calculated in Equation~\ref{eq:loss_classification_pixel}, where $i_{est}$ is $\prob(Y = fg)$, $(1-i_{est})$ is $\prob(Y = bg)$ and $i_{gt}$ is the ground truth value from the mask image.

\begin{equation} \label{eq:loss_classification_pixel}
    \begin{split}
        l^n (I_{est}^n, I_{gt}^n) = 
        & -w_{fg} i_{est} \log{(i_{gt})} \\
        & - w_{bg}(1-i_{est})\log{(1-i_{gt})}
    \end{split}
\end{equation}

Then, it is used as an input to calculate normalised loss for the whole image $\mathcal{L}_{mask}$ in Equation~\ref{eq:loss_classification_full}. A normalisation factor $\mathcal{N}$, which is a number of pixels in the image, allows us to keep the learning rate fixed, despite the variance of the input image size. 
    
\begin{equation}  \label{eq:loss_classification_full}
    \mathcal{L}_{mask} (I_{est}, I_{gt}) = \frac{1}{\mathcal{N}} \sum\limits_{n} l^n (i_{est},  i_{gt})
\end{equation}

Loss functions for both robot base coordinates and the coordinates of the robot joints are formulated as regression tasks. Both of them use Euclidean distance between estimated and ground truth values. Loss function for the 3D coordinates of robot joints $\mathcal{L}_{Jcoords}$ is described in Equation~\ref{eq:loss_joints_coords}, where $N_j$ is the number of joints, $J_{i}$ defines ground truth position of each joint and $E_{i}$ is the estimated position of each joint by the CNN.

\begin{equation} \label{eq:loss_joints_coords}
    \mathcal{L}_{Jcoords} = \frac{1}{N_j} \sum\limits_{i=1}^{N_j} \norm{J_{i}-E_{i}}_2
\end{equation}

Similarly, the loss function for the coordinates of the robot base $\mathcal{L}_{Bcoords}$ is shown in Equation~\ref{eq:loss_base_coords}. $B_{xyz}$ is the ground truth position of the robot base in 3D and $E_{xyz}$ is the estimated 3D position of the robot base. These positions are relative to the camera. Considering the goal of detecting the position of the robotic manipulator, estimating just Cartesian coordinates is sufficient. If necessary, the angles of each joint in relation to the robot base could be calculated by using coordinate positions.

\begin{equation} \label{eq:loss_base_coords}
    \mathcal{L}_{Bcoords} = \norm{B_{xyz}-E_{xyz}}_2
\end{equation}

The loss function to identify the robot type was defined as a categorical cross-entropy problem. It is commonly used for multi-class classification problems. $\mathcal{L}_{type}$ is calculated in Equation~\ref{eq:loss_robot_type}, where $p$ is the ground truth labels, $q$ are the predicted labels and $c\in R$, where $R$ are all the available types of robots in the dataset.

\begin{equation} \label{eq:loss_robot_type}
    \mathcal{L}_{type} = -\sum\limits_{c} p(c) \log{q(c)}
\end{equation}

The final loss function $\mathcal{L}_{final}$  is a weighted combination of all four previously described functions. The larger the weight $W$, the higher the emphasis on the correct prediction of the corresponding value. And the weights should be selected to have a good overall performance of the system. The calculation of $\mathcal{L}_{final}$ is described in Equation~\ref{eq:final_loss}.

\begin{equation}
\label{eq:final_loss}
    \begin{split}
        \mathcal{L}_{final} = 
        & W_{mask}\mathcal{L}_{mask} + W_{Jcoords}\mathcal{L}_{Jcoords} \\
        &+ W_{Bcoords}\mathcal{L}_{Bcoords} + W_{type}\mathcal{L}_{type}
    \end{split}
\end{equation}

In order to keep the CNN easily adaptable to other types of robots in the future, no prior information about the robot model is incorporated in the system. The raw CNN output is used to evaluate the accuracy of the results.

\section{CNN TRAINING}
\label{sec:training}

\begin{figure*}[ht]
\vspace{0.2cm}
    \centering
    \includegraphics[width=0.99\linewidth]{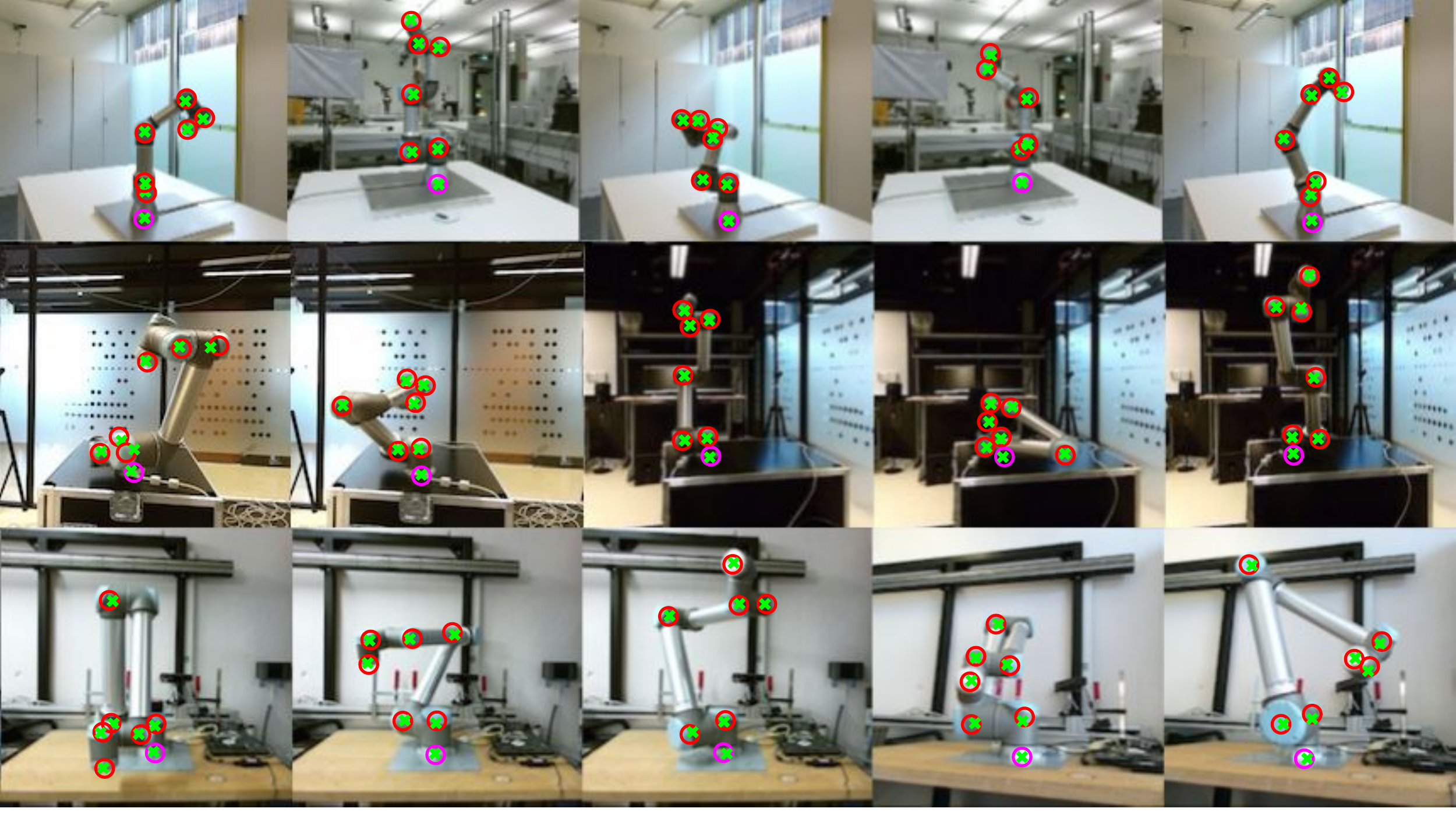}
    \caption{Estimated robot joint position coordinates marked on the images taken from the dataset. Due to difficulty in visualising 3D coordinates on printed figures, the estimated joint coordinates were mapped back into 2D images. Green crosses indicate the ground truth position, red circles indicate predicted positions of joints and magenta circles indicate the predicted position for the robot base.}
    \label{fig:marked_coordinates}
\vspace{-0.6cm}
\end{figure*}

Training of the multi-objective CNN is done for all four objectives at the same time. One possibility to adjust the quality of results is to adjust the weights given to the loss functions of each of the objectives when defining the final loss function of the system. In our case, the weight values were hand-selected using trial and error during the testing phase. Selected weight values were the following:
\begin{itemize}
    \item $W_{mask}$: $1.0$
    \item $W_{Jcoords}$: $1.5$
    \item $W_{Bcoords}$: $1.5$
    \item $W_{type}$: $0.3$
\end{itemize}

The training is done on the training set, including images of all three types of robots simultaneously. In total $926$ samples for UR3, $2520$ samples for UR5 and $904$ samples for UR10 were used

In order to speed up the process and have reasonably sized mini-batches, the input size of the images was reduced by half from the original dimensions, down to $256\times212$ pixels. The pixel intensity values of the input images were normalised to the range between 0 and 1. Furthermore, pixel values of the ground truth images are clipped to avoid division by zero in cases when the estimated mask fits the ground truth perfectly.
In order to avoid any training biases, the data were randomly shuffled and split into mini-batches of $64$ images each, fully utilising the memory of the GPU. The learning rate was set to $0.001$ at the beginning of the training and then gradually decreased to $0.000001$ as the training progressed. The CNN converged after $8000$ iterations. It took 60 hours to train the system using a regular NVIDIA GeForce 1080 GTX graphics card.

\section{RESULTS}
\label{sec:results}


The evaluation was done by testing our network on the test set and comparing the output against the ground truth data. The robot mask accuracy is defined by comparing a number of pixels in the CNN output image that match the ground truth mask. For the robot joint and base coordinates, Euclidean distance between the CNN estimated results and ground truth results was calculated. Robot type accuracy was computed by counting the percentage of correct classification instances. We compare the results against our previously presented C-CNN approach~\cite{2018arXiv180102025M}.

Robot mask classification achieved an accuracy of $98\%$, which is almost $3\%$ improvement compared to our previous method, as seen in Figure~\ref{fig:results_all}. A significant amount of this error comes from failing to estimate sharp corners in the mask image because CNN outputs slightly blurry mask compared to ground truth. It is likely that some post-processing would allow even further improvement by increasing the sharpness of the mask.

The overall error of the 3D position of robot joints was $3.16$ cm, which is a slight improvement compared to the error of $3.32$ cm in our previous work. If we analyse each joint separately, we can see the tendency of the joints closest to the robot base having a smaller average error, as well as smaller scatter, compared to the joints closer to the end-effector. Results are showing that in Figure~\ref{fig:results_joints}. It can be explained by analysing the reachability from the base of each of the joint. The end effector has the largest range of motion, and it reduces for the joints closer to the robot base. This means the range of possible positions varies significantly, and estimation is more difficult in the larger range of possible positions. However, the error difference is minor.

\begin{table}[h]
\caption{Summary of the results on the test set of a Multi-Objective CNN with a comparison to our previous work using C-CNN.}
\label{table:results_summary}
\centering
\begin{tabular}{ |p{3.2cm}||p{1.8cm}|p{1.8cm}|}
\hline
Measure & Current Work  & Previous Work \\
 \hline
 Mask Accuracy, \% & \boldmath$98\%$ & $94.6\%$ \\
 Robot Type Accuracy, \%  & \boldmath$98.3\%$ & --- \\
 Joint Pos Error (Mean) & \boldmath$3.16 cm$ & $3.32 cm$ \\
 Base Pos Error (Mean) & \boldmath$2.74 cm$ & $2.97 cm$ \\
 \hline
\end{tabular}
\vspace{-0.3cm}
\end{table}

The estimation of the position of the robot base in relation to the camera had an average error of $2.74$ cm. Once again, this is lower compared to C-CNN approach, where the same estimation error was $2.97$ cm. Robot type classification made just a few wrong decisions resulting in $98.3\%$ accuracy. The forward propagation time (detection speed) of the neural network was on average $15$ ms for one image, making it suitable for real-time applications.

The final results are summarised in Table~\ref{table:results_summary}, and the estimated coordinates by the full system marked over the dataset images can be seen in Figure~\ref{fig:marked_coordinates}. Because it is difficult to show 3D estimations on 2D figures, the visualisation of estimation is done by mapping the estimated 3D coordinates back onto input images.

Both in the current multi-objective CNN approach and the C-CNN method, we used exactly the same datasets for training and testing, so the results can be compared directly. Given the lack of similar work, no suitable benchmark was found to allowing a direct comparison of achieved results.

\section{CONCLUSIONS AND FUTURE WORK}
\label{sec:conclusion}

In this paper, we have presented a solution for detecting a robot manipulator and estimating the positions of its joints in a 2D camera image. A camera can be placed in arbitrary positions overlooking the robot workspace and the method successfully localizes the robot without the need for any additional setup or Hand-Eye calibration. This provides more flexible and quickly reconfigurable environment aware robotic setups for tasks like human-robot or robot-robot interaction. We have used three types of robots produced by Universal Robot for training and testing of the system: UR3, UR5 and UR10.

Our system uses a multi-objective convolutional neural network approach to achieve the goal. It optimises the system for four objectives simultaneously provides the mask of an area where the robot is present in the camera image, its base position in relation to the camera, 3D positions of the joints of the robot as well as the type of the robot, respectively the 3D joint position error was less than $3.16$ cm, the robot mask accuracy was $98\%$ and the robot type was successfully recognised in over $98.3\%$ of cases. These results are an improvement of our previously presented C-CNN approach, both in accuracy and flexibility of the system.

Given current results, the continuation of work will be to apply this method in more complex environments containing multiple robots and people working in the same workspace. Self-occlusions were present in the tested datasets and some minor occlusions of other objects, however, more evaluation is needed using cases like people or other machinery passing by between the camera and the robot blocking the view.

This work has multiple possible applications. One would be the safety aspect of identifying robots in robotised environments like factory floors, warehouses or automated surgery rooms where an operator has a wearable camera detecting robots in the field of view. Another application would be for robot-robot interaction. With swarm robotics, both homogeneous and heterogeneous, and different sizes, direct communication between them is not always reliable. Our approach would allow the robots to observe and track each other using small cameras and identify the intentions of other robots in the surroundings.



For the human-robot collaboration tasks, a person tracking can be achieved using devices like Leap Motion or skeleton tracking to estimate of the relative hand positions to the robot. This can be used for tool handover between the person and the robot, working towards a common goal or even hand-gesture control, while avoiding any unwanted physical contact between the two.

In the future, we plan to test the system with more types of the robots by using transfer learning on pre-trained CNN. In this case, the dataset needed to teach to identify a new robot type should be significantly reduced compared to the current setup. Adding human skeleton tracking would move the work closer to the real-world human-robot interaction tasks. The system will be tested in some use case scenarios to identify the robustness in less controlled environments with more illumination changes and changing setups.

\section*{ACKNOWLEDGMENT}
This work is partially supported by The Research Council of Norway as a part of the Engineering Predictability with Embodied Cognition (EPEC) project, under grant agreement 240862, and by the Austrian Ministry for Transport, Innovation and Technology (BMVIT) within the project framework CollRob (Collaborative Robotics).

\addtolength{\textheight}{-12cm}   






\bibliographystyle{IEEEtran}
\bibliography{IEEEexample}

\end{document}